\documentclass[10pt,twocolumn,letterpaper]{article}

\usepackage{iccv}
\usepackage{times}
\usepackage{epsfig}
\usepackage{graphicx}
\usepackage{amsmath}
\usepackage{amssymb}

\usepackage{stackengine} 
\usepackage{dblfloatfix} 
\usepackage{subcaption}
\usepackage{pifont}
\usepackage{mathtools}
\usepackage{algorithm}
\usepackage[noend]{algorithmic}
\usepackage[dvipsnames]{xcolor}
\newcommand\oast{\stackMath\mathbin{\stackinset{c}{0ex}{c}{0ex}{\ast}{\bigcirc}}}
\newcommand{\norm}[1]{\left\lVert#1\right\rVert}
\newcommand{\xmark}{\ding{55}}

\newcommand{\mytensor}[1]{\ensuremath{\mathcal{#1}}}
\newcommand{\myvector}[1]{\ensuremath{\mathbf{#1}}}
\newcommand{\mymatrix}[1]{\ensuremath{\mathbf{#1}}}
\newcommand{\mycolon}{\,\boldsymbol{:}\,}

\newcommand{\myst}{\ensuremath{^{\text{st}}}\,}
\newcommand{\mynd}{\ensuremath{^{\text{nd}}}\,}

\usepackage[pagebackref=true,breaklinks=true,letterpaper=true,colorlinks,bookmarks=false]{hyperref}

\iccvfinalcopy 

\ificcvfinal\fi
\begin{document}

\title{Matrix and tensor decompositions for training binary neural networks}

\author{Adrian Bulat \qquad Jean Kossaifi \qquad Georgios Tzimiropoulos \qquad Maja Pantic
        \vspace{5pt}\\
		Samsung AI Center, Cambridge\\
		United Kingdom\\
		{\tt\small \{adrian.bulat, j.kossaifi, georgios.t, maja.pantic\}@samsung.com}
		}

\maketitle

\begin{abstract}

   This paper is on improving the training of binary neural networks in which both activations and weights are binary. 
  While prior methods for neural network binarization binarize each filter independently,
  we propose to instead parametrize the weight tensor of each layer using matrix or tensor decomposition.
  The binarization process is then performed using this latent parametrization, via a quantization function (e.g. sign function) applied to the reconstructed weights.
  A key feature of our method is that while the reconstruction is binarized, the computation in the latent factorized space is done in the real domain. 
  This has several advantages:
  (i) the latent factorization enforces a coupling of the filters before binarization, which significantly improves the accuracy of the trained models. 
  (ii) while at training time, the binary weights of each convolutional layer are parametrized using real-valued matrix or tensor decomposition, during inference we simply use the reconstructed (binary) weights. 
  As a result, our method does not sacrifice any advantage of binary networks in terms of model compression and speeding-up inference. 
  As a further contribution, instead of computing the binary weight scaling factors analytically, as in prior work, we propose to learn them discriminatively via back-propagation. 
  Finally, we show that our approach significantly outperforms existing methods when tested on the challenging tasks of (a) human pose estimation (more than 4\% improvements) and (b) ImageNet classification (up to 5\% performance gains).

\end{abstract}

\section{Introduction}\label{sec:introduction}

One key aspect of the performance of deep neural networks is the availability of abundant computational resources (i.e high-end GPUs) during both training and inference. However, often, such models need to be deployed on devices with limited resources such as smartphones, FPGAs or embedded boards. To this end, there is a plethora of works that attempt to miniaturize the models and speed-up inference with popular directions including matrix and tensor decomposition~\cite{lebedev2014speeding,kim2015compression}, weights pruning~\cite{han2015learning} or network quantization~\cite{courbariaux2014training,lin2015fixed}. Of particular interest in this work is the extreme case of quantization -- binarization, where all the weights and features are restricted to 2 states only. Such networks can achieve a compression rate of up to $32\times$ and an even higher order speed-up that can go up to $58\times$~\cite{rastegari2016xnor,courbariaux2015binaryconnect}. 
Despite these attractive properties, training binary networks to a comparable accuracy to that of their real-valued counterparts is still an open problem. For example, there is $ \sim 20 \%$ accuracy drop between real and binary networtks on ImageNet~\cite{rastegari2016xnor}, and $ \sim 9 \%$ difference for human pose estimation on MPII \cite{bulat2017binarized}.    

Most current works that attempt to improve the accuracy of binary network fall into two broad categories: a) methodological changes and b) architectural improvements. The authors of~\cite{courbariaux2015binaryconnect} propose to binarize the weights using the $\text{sign}(x)$ function, with encouraging results on a few selected datasets. Because the representational power of binary networks is very limited, the authors of~\cite{rastegari2016xnor} propose to add a scaling factor to the weights and channels of each convolutional layer, showing for the first time competitive results on ImageNet. From an architectural point of view, the method of~\cite{bulat2017binarized} proposes a novel residual module specially designed for binary networks, while in~\cite{tang2018quantized}, the authors incorporate  densenet-like connections into the U-Net architecture.

In this work, we propose a simple method to improve the accuracy of binary networks by introducing a linear or multi-linear re-parametrization of the weight tensor during training. Let's consider a \(4\)--dimensional weight tensor \(\mytensor{W}\in\mathbb{R}^{O \times C \times w \times h}\).
A common limitation in prior work is that each filter $\mytensor{W}_i\in\mathbb{R}^{C\times w \times h}$ (a slice of \(\mytensor{W}\)) of a given convolutional layer is binarized independently as follows: 
\begin{equation}
    \mytensor{B}_i = \text{sign}(\mytensor{W}_i). \nonumber
\end{equation} 
In contrast, our key idea in this work is to model the filters jointly by re-parametrizing them in a shared subspace using a matrix or tensor decomposition, and then binarizing the weights. A simplified version of our idea can be described as follows:
\begin{eqnarray}
    \mytensor{W} & = & \mymatrix{U}\mymatrix{V}\nonumber \\
    \mytensor{B}_i & = & \text{sign}(\mytensor{W}_i).\nonumber
\end{eqnarray}
This allows us to introduce an inter-dependency between the to-be-binarized weights through the shared factor $\mymatrix{U}$ either at a layer level or even more globally at a network level. A key feature of our method is that the decomposition factors (i.e $\mymatrix{U}, \mymatrix{V}$) are kept real during training. This allows us to introduce additional redundancy which, as we will show facilitates learning. 

Note that this latent parametrization is used only during training. During inference, our method only uses the reconstructed weights, which have been binarized using the sign function (the decomposition factors are neither used nor stored). Hence, our method does not sacrifice any of the advantages of binary networks in terms of model compression and inference speed-up.

In summary, we make the following \textbf{contributions}:
\begin{itemize}
    \item We are the first to propose parameterizing the binarized weights of a neural networks using a real-valued linear and multi-linear decomposition (at training time). In other words, we enforce a shared subspace between the filters of the convolutions, as opposed to prior work that model and binarize each filter independently. This novel approach allows us to further improve the accuracy of binary networks
    without sacrificing any of their advantage in terms of model compression and speeded-up inference.
    (Section~\ref{ssec:binary-tensorized-convolutional}).
    
    \item We revise the convolutional approximation proposed in~\cite{rastegari2016xnor}: $(\mytensor{I} * \mytensor{W} = (\text{sign}(\mytensor{I}) * \text{sign}(\mytensor{W}))\alpha $, where the scaling factor $\alpha$ is computed analytically from  $\alpha = \frac{1}{n}\norm{\mytensor{W}}_{l1}$ and propose to instead learn it discriminatively via back-propagation at train time (Section~\ref{ssec:learnable-scaling}).
    
    \item We explore several types of decomposition (SVD and Tucker) applied either layer-by-layer or jointly to the entire network as a whole (Section~\ref{sec:method}). We perform in-depth ablation studies that help shed light on the advantages of the newly proposed method.
    
    \item We show that our method significantly advances the state-of-the-art for two important computer vision tasks: human pose estimation on MPII and large-scale image classification on ImageNet (Section~\ref{sec:experiments}).
\end{itemize} \section{Related work}\label{sec:related-work}
In this section, we review the related work, in terms of neural network architectures~(\ref{ssec:efficient-cnns}), network binarization~(\ref{ssec:network-binarization}) and tensor methods~(\ref{ssec:tensor-methods}).

\subsection{Efficient neural network architectures}\label{ssec:efficient-cnns}
 Despite the remarkable accuracy of deep neural networks on a large variety of tasks, deploying such networks on devices with low computational resources is highly impractical. Recently, a series of works have attempted to alleviate this issue via architectural changes applied either at the block or architecture level.

\textbf{Block-level optimization.} In~\cite{he2016deep} He et al. proposes the so-called bottleneck block that attempts to reduce the number of $3\times3$ filters using 2 convolutional layers with a $1\times1$ kernel that project the features into a lower dimensional subspace and back.  The authors from ~\cite{xie2016aggregated} introduce a new convolutional block that splits the module into a series of parallel sub-blocks with the same topology. The resulting block has a smaller footprint and higher representational power. In a similar fashion, MobileNet~\cite{howard2017mobilenets} and its improvement~\cite{sandler2018mobilenetv2} make use of depth-wise convolutional layers with the later proposing an inverted bottleneck module. In~\cite{zhang2018shufflenet}, the authors combine point-wise group convolution and channel shuffle incorporating them in the bottleneck structure.

Note, that in this work we do not attempt to improve the architecture itself and simply use the already well-established basic block with pre-activation introduced in~\cite{he2016identity} (see Fig.~\ref{fig:basic-block}).

\textbf{Network-level optimization.} The Dense-Net architecture~\cite{huang2016densely} proposes to inter-connect each layer to every other layer in a feed-forward fashion. This results in a better gradient flow and higher performance per number of parameters ratio. Variations of it were later adopted for other tasks, such as human pose estimation~\cite{tang2018quantized}. In~\cite{redmon2016you} and its follow-up~\cite{redmon2017yolo9000} the authors introduce the so-called YOLO architecture which proposes a new framework for object detection and an optimized architecture for the network backbone that can run real-time on a high-end GPU.

\subsection{Network binarization}\label{ssec:network-binarization}
Another direction for speeding-up neural networks is network quantization. This process reduces the number of possible states that the weights and/or the features can take and has become increasingly popular with the advent of low-precision computational hardware. 

While normally CNNs operate using float-32 values, the work of~\cite{courbariaux2014training, lin2015fixed} proposes to use 16-- and 8-bit quantization showing in the process insignificant performance drop on a series of small datasets (MNIST, CIFAR10). Zhou et al. ~\cite{zhou2016dorefa} proposes to allocate a different numbers of bits for the network parameters (1 bit), activations (2 bits) and gradients (6 bits), the values of which are selected based on their sensitivity to numerical inaccuracies. \cite{wang2018two}~propose a two-step n-bits quantization ($n\geq2$), where the first step consists of learning a low-bit code and the second in learning a transformation function. In~\cite{faraone2018syq}, the authors propose to learn a 1-2 bit quantization for the weights and 2-8 for activations by learning a symmetric codebook for each particular weights subgroup. While such methods can lead to significant space and speed gains, the most interesting case is that of binarized neural networks. Such networks have their features and weights quantized to two states only. In~\cite{soudry2014expectation} the authors propose to binarize the weights using the $sign$ function. Follow-up work~\cite{courbariaux2015binaryconnect, courbariaux2016binarized} further improve these results, by binarizing both the activations and the weights. In such type of networks the multiplications inside the convolutional layer can be replaced with XOR bitwise operations. The current state-of-the-art binarization technique is the XNOR-Net method~\cite{rastegari2016xnor} that proposes a real-valued scaling factor for the weights and inputs. The proposed XNOR-Net method~\cite{rastegari2016xnor} is the first to report good results on a large scale dataset (ImageNet). In~\cite{bulat2017binarized}, the authors propose a new module specifically designed for binary networks. The work of ~\cite{mishra2017wrpn} explores ways of increase the quantized network accuracy by increasing its width (i.e number of channels) motivated by the idea that often the activations are taking most of the memory during training. In a similar fashion, in~\cite{lin2017towards} the authors use up to 5 parallel binary convolutional layers to approximate a real one, as such increasing the size and computational requirements of the network up to $5\times$. \cite{zhou2018explicit}~proposes a loss-aware binarization method that jointly regularizes the weights approximation error and the accompanying loss, however this method quantizes the weights while leaving the features real. ~\cite{hu2018training} proposes a semi-binary decomposition of the binary weight tensor into two binary matrices and a diagonal real-valued one which are used (instead of the actual binary weights) during test time. As mentioned by the authors the proposed binary-to-(semi-)binary decomposition is a difficult optimization problem and hence harder to train. More importantly, and in contrary to our method, in this approach, the activations are kept real. 

In this work, we propose to improve the binarization process itself, introducing a novel approach that increases the representation power and flexibility of binary weights at train time via matrix and tensor re-parametrization while maintaining the same structure and very large speed gains during inference.

\subsection{Tensor methods}\label{ssec:tensor-methods}
Tensor methods offer a natural extension of the more traditional algebraic methods to higher orders that naturally arise in convolutional networks. As such, this family of methods is actively deployed, both for compressing and speeding-up the networks via re-parametrization~\cite{lebedev2014speeding,yong2015compression,astrid2017cp,yong2015compression}, or by taking advantage directly of the higher order dimensionality present in the data~\cite{kossaifi_tcl,kossaifi2018tensor}.

Separable convolutions, recently popularized in~\cite{chollet2017xception}, are one such example that can be obtained by applying a CP decomposition to the layer weights. In \cite{lebedev2014speeding}, the weights of each convolutional layer are decomposed into a sum of rank--\(1\) tensors using a CP decomposition in an attempt to speed-up the convolutional modules. At inference time this is achieved by replacing the original layers with a set of smaller ones where the weights of each newly introduced layer represent the factors themselves.
Similarly, in~\cite{yong2015compression} the authors re-parametrize the layer weights using a Tucker decomposition. At test time, the resulting module resembles a bottleneck~\cite{he2016deep} block. \cite{tai2015convolutional} propose to decrease the redundancy typically present in large neural networks by  expressing each layer as the composition of two convolutional layers with less parameters. Each 2D filter is approximated by a sum of rank--\(1\) tensors. However, this can be applied only for convolutional layer which have a kernel size larger than 1. While most of the works mentioned above are applied to convolutional layers other types can be parameterized too. In all the aforementioned works, tensor decompositions are applied to individual convolutional layers. More recently, the work of~\cite{kossaifi2018parametrizing} proposed a simple method for whole network tensorization using a single high-order tensor. 

To our knowledge, none of the above methods have been applied to binary networks. By doing so, our approach allows us to combine the best of both words: take advantage of the very high compression rate and speed-up typically offered by binarized networks while maintaining the increased representational power offered by the tensor re-parametrization methods. A crucial aspect of this re-parametrization is that it enables us to enforce an inter-dependency between the binary filters, which were previously treated independently by prior work on binarization~\cite{rastegari2016xnor,courbariaux2016binarized}.

\subsection{Human pose estimation}\label{ssec:human-pose-estimation}
While a complete review of recent work on human pose estimation goes beyond the scope of this paper, the current state-of-the-art on single person human pose estimation is based on the so-called ''Hourglass``(HG) architecture~\cite{newell2016stacked} and its variants~\cite{bulat2016human,tang2018deeply,ke2018multi,yang2017learning}. Most of this prior work focuses on achieving the highest performance without imposing any computational restrictions. Only recently, the work in~\cite{tang2018quantized} and ~\cite{bulat2017binarized} study this problem in the context of quantized neural networks. In~\cite{tang2018quantized} the authors propose an improved HG architecture that makes use of dense connections~\cite{huang2016densely} while~\cite{bulat2017binarized} introduces a novel residual block specially tailored to binarized neural networks. In contrast with the aforementioned methods, in this work, instead of improving the network architecture itself, we propose a novel, improved binarization technique that is independent of the network and task at hand.

\begin{figure}
    \centering
    \includegraphics[width=0.75\linewidth,trim={0.5cm 0.5cm 0.5cm 0.5cm},clip]{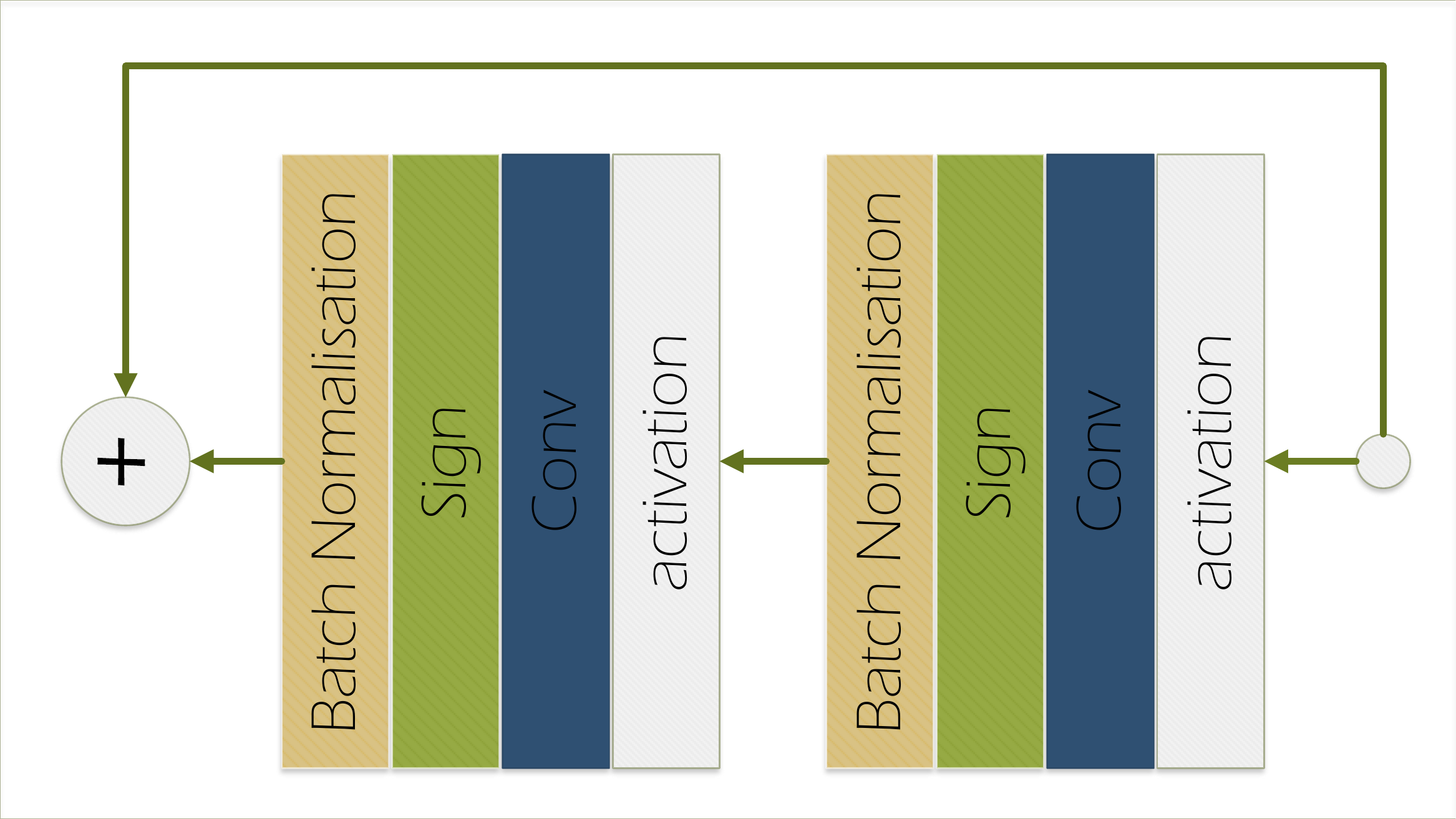}
    \caption{The original residual basic block proposed in~\cite{he2016deep} with the changes introduced ~\cite{rastegari2016xnor}.}
    \label{fig:basic-block}
\end{figure}
 \section{Background}\label{ssec:background}

Let $\mytensor{W}\in \mathbb{R}^{O\times C\times w \times h}$ and $\mytensor{I}\in \mathbb{R}^{C\times w_{in} \times h_{in}}$ denote the weights and respectively, the input of the $L$-th convolutional layer, where $O$ represents the number of output channels, $C$ the number of input channels and ($w,h$) the width and height of the convolutional kernel. $w_{in}\leq w$ and $h_{in}\leq h$ represent the spatial dimension of the input features $\mytensor{I}$. In its simplest form the binarization process can be achieved by taking the sign of the weights and respectively, of the input features where:

\begin{equation}\label{eq:sign}
    \text{sign}(x) = \begin{cases} -1, & \mbox{if } x\leq 0 \\ 1, & \mbox{if } x>0 \end{cases}
\end{equation}

However, such approach leads to sub par performance on the more challenging datasets. In~\cite{rastegari2016xnor}, Rastegari et al. proposes to alleviate this by introducing a real-valued scaling factor that boosts the representational power of such networks:

\begin{equation}\label{eq:binarize}
    \mytensor{I} \ast \mytensor{W} = \left(\text{sign}(\mytensor{I}) \oast \text{sign}(\mytensor{W}) \right) \odot \myvector{\alpha},
\end{equation}
where $\myvector{\alpha}_{i} = \frac{1}{n}\norm{\mytensor{W}(i, \mycolon, \mycolon, \mycolon)}_{\ell 1}, i=\{1, 2, \cdots, O\}$ 
and $n = C\times w \times h$.We denotes as $\ast$ the real-valued convolutional operation and $\oast$ its binary counterpart, implemented using XNOR bitwise operations. Note, that while in~\cite{rastegari2016xnor} a scaling factor is proposed for both input features and weights, in this work we use only the later since removing the first significantly speeds-up the network at a negligible drop in accuracy~\cite{rastegari2016xnor,bulat2017binarized}.

\section{Method}\label{sec:method}

In this section, we present our novel binarization method that aims to increase the representational power of the binary networks by enforcing for the first time an inter-dependency between the binary filters via a linear or multi-linear over-parametrization of the weights. We start by introducing some necessary notation (Section~\ref{ssec:notation}). We then continue by describing the main algorithm and its variations (Section~\ref{ssec:binary-tensorized-convolutional}). Finally, in Section~\ref{ssec:learnable-scaling} we describe how to further improve the proposed binarization technique by optimizing the scaling factor with respect to the target loss function via back-propagation. 

\subsection{Notation}\label{ssec:notation}
We denote vectors (1\myst order tensors) as \(\myvector{v}\), matrices (2\mynd order tensors) as \(\mymatrix{M}\) and tensors of order $\geq 3$, as \(\mytensor{X}\). We denote element \((i_0, i_1, \cdots, i_N)\) of a tensor as \(\mytensor{X}_{i_0, i_1, \cdots, i_N} \,\text{ or } \, \mytensor{X}(i_0, i_1, \cdots, i_N)\). A colon is used to denote all elements of a mode, e.g. the mode-1 elements of \(\mytensor{X}\) are denoted as \(\mytensor{X}(:, i_2, i_3, \cdots, i_N)\).

\noindent \textbf{Tensor contraction:} we define the n--mode product (contraction between a tensor and a matrix), for a tensor \(\mytensor{X} \in \mathbb{R}^{D_0 \times D_1 \times \cdots \times D_N}\) and a matrix \( \mymatrix{M} \in \mathbb{R}^{R \times D_n} \), as the tensor \(\mytensor{T} = \mytensor{X} \times_n \mymatrix{M} \in \mathbb{R}^{D_0 \times \cdots \times D_{n-1} \times R \times D_{n+1} \times \cdots \times D_N} \), with:
\begin{equation}
    \mytensor{T}_{i_0, i_1, \cdots, i_n} = \sum_{k=0}^{D_n} \mymatrix{M}_{i_n, k} \mytensor{X}_{i_0, i_1, \cdots, i_n}.
\end{equation}

\subsection{Matrix and tensor re-parametrizations for training binary CNNs}\label{ssec:binary-tensorized-convolutional}

A key limitation of prior work on training binary networks is that each of the filters of the weight tensor in each convolutional layer is binarized independently, without imposing any relation between the filters explicitly. To alleviate this, we propose to increase the representational power of the binary network via re-parametrization. During training, we propose to express the to-be-binarized weights~$\mytensor{W}$ of each convolutional layer using a linear or multi-linear \textit{real-valued} decomposition:  
\begin{equation}
    \mytensor{W} = \textbf{ReconstructWeights}(\mymatrix{\Theta_0}, \mymatrix{\Theta_{1}}, ... , \mymatrix{\Theta_{N}}),
\end{equation}
where the function \textbf{ReconstructWeights}(;) is specific to the decomposition used, there exists at least one decomposition factor $\mymatrix{\Theta_k}$ which is shared among all filters in $\mytensor{W}$, and the set of all decomposition factors $\mymatrix{\Theta_i}, i=1, \dots, N$ are all real-valued. Using a real-valued decomposition is a key feature of the proposed approach as it allows us to introduce additional redundancy which as we show facilitates learning. 

Note that when training is done, our method simply uses the  reconstructed weights which are converted to binary numbers using the sign function. Hence, during inference the factors $\mymatrix{\Theta_i}, i=1, \dots, N$ are neither used nor need to be stored, only the reconstructed binarized weights are used. Hence, our method does not sacrifice any advantage of binary networks in terms of model compression and speeding-up inference.

In the context of this work, we explore two different decompositions: SVD and Tucker. We apply these decompositions in two different ways: \textit{layer-wise} and \textit{holistically}. Layer-wise decomposition refers to modeling the weight tensor of each convolutional layer separately, i.e. performing a different decomposition for each convolutional layer (e.g. \cite{lebedev2014speeding, yong2015compression, tai2015convolutional}). We note that this is the standard way in literature that SVD and Tucker decomposition have been applied for neural network re-parametrization.  More recently, the work of~\cite{kossaifi2018parametrizing} proposed a single method for whole network tensorization using a single high-order tensor. We refer to this tensorization approach as holistic.

Note, that unlike other binarization methods where two set of weights are explicitly stored in memory and swapped at each iteration~\cite{rastegari2016xnor} our method can deal with this implicitly without a secondary copy. This is due to the fact that the factors are always real-valued and are reconstructed and binarized on-demand during training.

The entire proposed method for binarization is described in Algorithm \ref{alg:trainbinconv}:

\renewcommand{\algorithmicrequire}{\textbf{Input:}}
\renewcommand{\algorithmicensure}{\textbf{Output:}}
\begin{algorithm}[thb]	
{
  \caption{Training an $L$-layer CNN with binary weights via matrix or tensor decomposition. The rows colored in {\color{blue} blue} are the changes introduced by our method when compared against the approach proposed in~\cite{rastegari2016xnor}.}
  \label{alg:trainbinconv}       
  \begin{algorithmic}[1]
      \REQUIRE A minibatch of inputs and targets  ($\mytensor{I}, \mytensor{Y}$), cost function $C(\mytensor{Y},\hat{\mytensor{Y}})$, current set of matrices from which the weights can be reconstructed  $\mymatrix{\Theta_{0}}^t, \mymatrix{\Theta_1}^t, \cdots, \mymatrix{\Theta_{N}}^t$ (obtained using one of the methods described in Sections~\ref{sssec:layer-wise-svd}-\ref{sssec:holistic-tucker}) and current learning rate $\eta^t$.  Optionally, if the scaling factor is computed using the method described in Section~\ref{ssec:learnable-scaling}, the current weights scales $\myvector{\alpha}^t$.
  \ENSURE updated factors $\mymatrix{\Theta_{0}}^{t+1}, \mymatrix{\Theta_{1}}^{t+1}, \cdots, \mymatrix{\Theta_{N}}^{t+1}$ and updated learning rate $\eta^{t+1}$.  If $\myvector{\alpha}$ is computed using the method from Section~\ref{ssec:learnable-scaling}, also return the updated $\myvector{\alpha}^{t+1}$.
  \STATE Binarizing weight filters:
  \FOR{$l=1$ to $L$}
      \STATE {\color{blue} $\mytensor{W}_l = \textbf{ReconstructWeights}(\mymatrix{\Theta_0^{l}}^t, \mymatrix{\Theta_{1}^{l}}^t, ... , \mymatrix{\Theta_{N}^{l}}^t)$ ~~//~ {\scriptsize Using Eqs.~\ref{eq:svd} or~\ref{eq:binarize-tucker} or~\ref{eq:holistic}}}
      \IF{ $\alpha$ is defined} 
          \STATE { \color{blue}$\mytensor{B}_{l}=\text{sign}(\mytensor{W}_{l}^t)$ }
          \STATE { \color{blue} $\widetilde{\mytensor{W}}_{l} = \alpha_{l}\mytensor{B}_{l}$ ~~//~ {\scriptsize If using method proposed in Section~\ref{ssec:learnable-scaling}}}
      \ELSE
          \FOR{$k^{\text{th}}$ filter in $l^{\text{th}}$ layer}
              \STATE $\alpha_{lk}=\frac{1}{n}\Vert\mytensor{W}_{lk}^t\Vert_{\ell 1}$
              \STATE $\mytensor{B}_{lk}=\text{sign}(\mytensor{W}_{lk}^t)$
              \STATE $\widetilde{\mytensor{W}}_{lk} = \alpha_{lk}\mytensor{B}_{lk}$
          \ENDFOR
      \ENDIF
   \ENDFOR
  \STATE $\hat{\mytensor{Y}}=$ ~~\textbf{ForwardPass}$(\mytensor{I},\mytensor{\widetilde{\mytensor{W}}})$~~//~{\scriptsize standard forward propagation where the convolutional operations use the reconstructed binarized weights $\widetilde{\mytensor{W}}$}
  
  \STATE $\frac{\partial C}{\partial \Theta} =$ \textbf{BackwardPass}$(\frac{\partial C}{\partial \hat{\textbf{Y}}}, \widetilde{\mathcal{W}})$ ~~//~{\scriptsize standard backward propagation where gradients are computed using the reconstructed binary weights with respect to the factors}
  \STATE Update the matrices using an update rule (i.e ADAM, RMSprop):
  {\color{blue} 
  \FOR{$i=0$ to $N$}
     \STATE { \color{blue} $\mymatrix{\Theta}_{i}^{t+1}=$ \textbf{UpdateParameters}$(\mymatrix{\Theta}_i^t,\frac{\partial C}{\partial \mymatrix{\Theta}_i},\eta_t)$ }
  \ENDFOR
  }
  {\color{blue} 
  \IF{ $\alpha$ is defined}
      \STATE { \color{blue} $\myvector{\alpha}^{t+1}=$ \textbf{UpdateParameters}$(\myvector{\alpha}^t,\frac{\partial C}{\partial \myvector{\alpha}},\eta_t)$ }
  \ENDIF
  }
 \STATE $\eta^{t+1}=$ \textbf{UpdateLearningRate}$(\eta^t, t)$
  \end{algorithmic}
  } \normalsize
\end{algorithm}

\subsubsection{Layer-wise SVD decomposition}\label{sssec:layer-wise-svd}
Let $\mymatrix{W}\in \mathbb{R}^{O \times (Cwh)}$ be the reshaped version of weight $\mytensor{W}$ of the $L$-th layer. By applying an SVD decomposition we can express $\mymatrix{W}$ as follows:

\begin{equation}\label{eq:svd}
    \text{sign}(\mymatrix{W}) = \text{sign}(\mymatrix{U} \mymatrix{\Sigma} \mymatrix{V^{T}}),
\end{equation}
where $\mymatrix{\Sigma} \in \mymatrix{R}^{M \times M}, M \leq \text{min}(O, Cwh)$ is a diagonal matrix and $\mymatrix{U}\in\mathbb{R}^{O \times K}$, $\mymatrix{V}\in\mathbb{R}^{K \times (Cwh)}$. By substituting eq.~(\ref{eq:svd}) in (\ref{eq:binarize}) we obtain:

\begin{equation}\label{eq:binarize-svd}
    \mathbf{I} \ast \mathbf{W} = (\text{sign}(\mathbf{I}) \oast \text{sign}(\mathbf{U} \mathrm{\Sigma} \mathbf{V^{T}})) \odot \mathbf{\alpha},
\end{equation}
where $\mathbf{U}$ and $\mathbf{V}$ are learned via backpropagation.

When evaluated on the validation set of MPII, re-parametrizing the weights layer-wise using SVD improves the performance by 0.3\% (see Table~\ref{tab:main-results-mpii}).

\subsubsection{Layer-wise Tucker decomposition}\label{sssec:layer-wise-tucker}

While the SVD decomposition shows some benefits on the MPII dataset, one of its core limitation is that it requires reshaping the weight to a 2D matrix, those losing the important spatial structure information present in them.

To alleviate this, we propose using the Tucker decomposition, a natural extension of SVD for higher order tensors. Using the Tucker decomposition we can express the binary weights as follow:

\begin{equation}\label{eq:binarize-tucker}
    \text{sign}(\mytensor{W}) = \text{sign}(\mytensor{G} \times_0 \mymatrix{U}^{(0)} \times_1 \mymatrix{U}^{(1)} \times_2 \mymatrix{U}^{(2)}  \times_3 \mymatrix{U}^{(3)}),
\end{equation}
where $\mytensor{G} \in \mathbb{R}^{O \times C \times w \times h} $ is a full-rank core and $(\mymatrix{U}^{(0)},\mymatrix{U}^{(1)},\mymatrix{U}^{(2)},\mymatrix{U}^{(3)})$ a set of factors.

The results from Table~\ref{tab:main-results-mpii} and~\ref{tab:main-results-imagenet} confirm the proposed hypothesis, showing an improvement of more than 0.7\% on top of the gains offered by the SVD decomposition. 

\begin{table}[!htbp]
\vspace{10pt}
\small
	\begin{center}
		\begin{tabular}{|c|c|c|c|}
			\hline
			Decomposition & Holistic & Learn. alpha & PCKh \\
			\hline\hline
			None & - & \xmark  & 78.4\% \\
			None & - & \checkmark  & 79.3\% \\
			SVD & \xmark & \xmark  & 78.7\% \\
			SVD & \xmark & \checkmark  & 79.0\% \\
			Tucker & \xmark & \xmark  & 79.3\% \\
			Tucker & \xmark & \checkmark  & 79.9\% \\
			Tucker & \checkmark & \xmark  & 82.0\% \\
			\textbf{Tucker} & \checkmark & \checkmark  & \textbf{82.5}\% \\
			
            \hline
		\end{tabular}
	\end{center}
	\caption{PCK-h based results on the validation set of MPII for different variations of the proposed binarization method. Notice that the proposed holistic approach significantly outperforms the baseline.}
	\label{tab:main-results-mpii}
\end{table}

\begin{figure*}[!htbp]
    \centering
    \begin{subfigure}[t]{0.24\textwidth}
    \centering
    \includegraphics[width=1.0\linewidth]{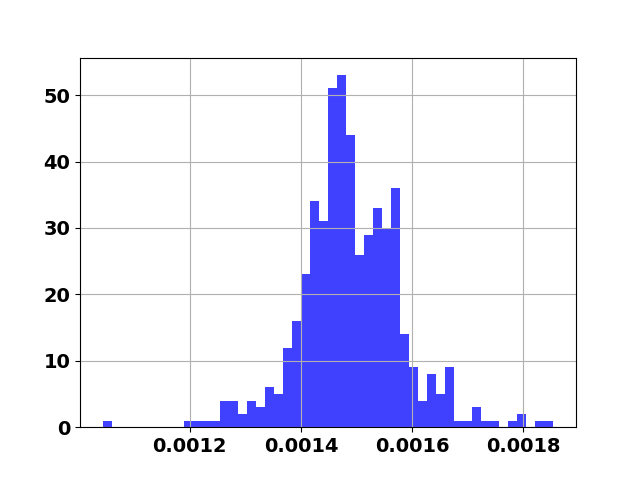}
    \end{subfigure}
    \hfill
    \begin{subfigure}[t]{0.24\textwidth}
    \centering
    \includegraphics[width=1.0\linewidth]{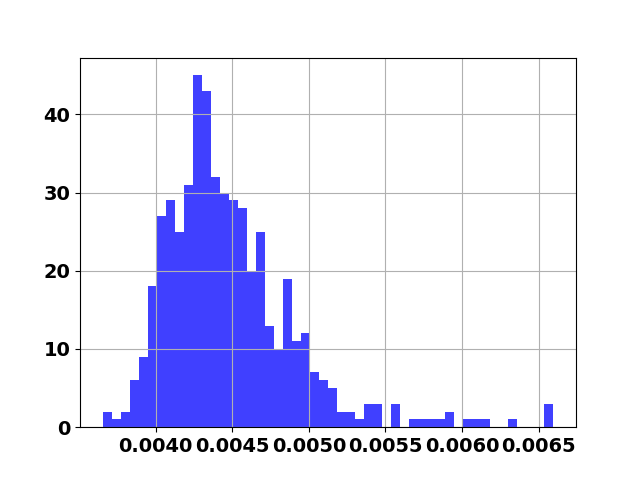}
    \end{subfigure}
    \hfill
    \begin{subfigure}[t]{0.24\textwidth}
    \centering
    \includegraphics[width=1.0\linewidth]{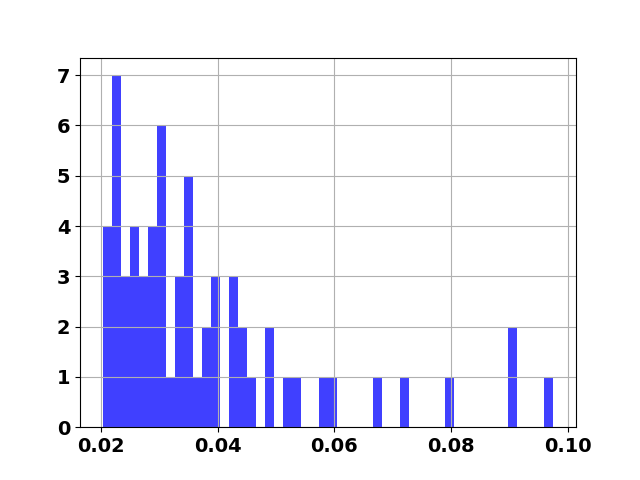}
    \end{subfigure}
    \hfill
    \begin{subfigure}[t]{0.24\textwidth}
    \centering
    \includegraphics[width=1.0\linewidth]{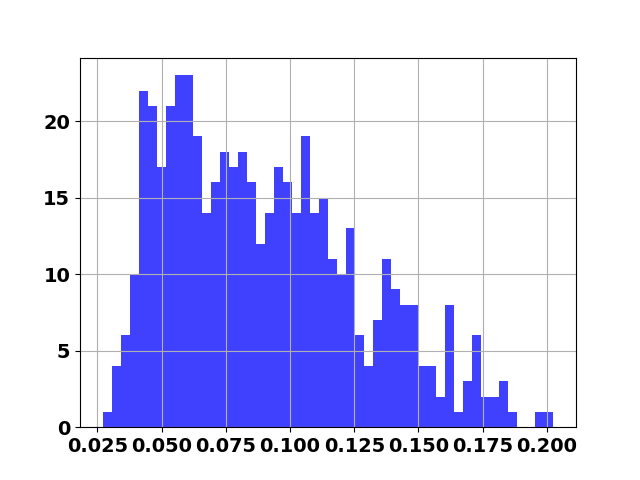}
    \end{subfigure}\\
    \begin{subfigure}[t]{0.24\textwidth}
    \centering
    \includegraphics[width=1.0\linewidth]{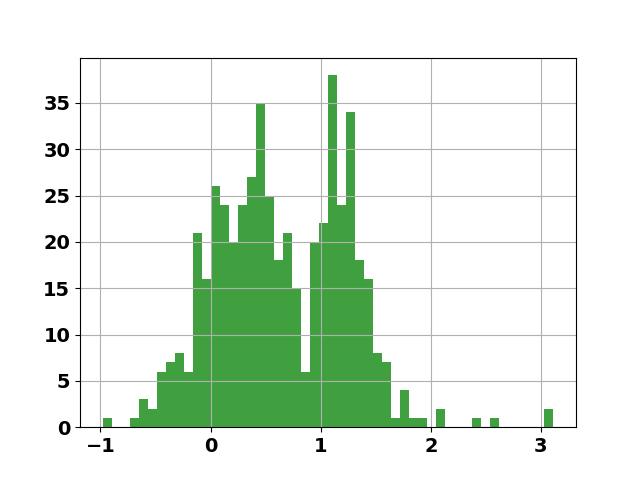}
    \end{subfigure}
     \hfill
    \begin{subfigure}[t]{0.24\textwidth}
    \centering
    \includegraphics[width=1.0\linewidth]{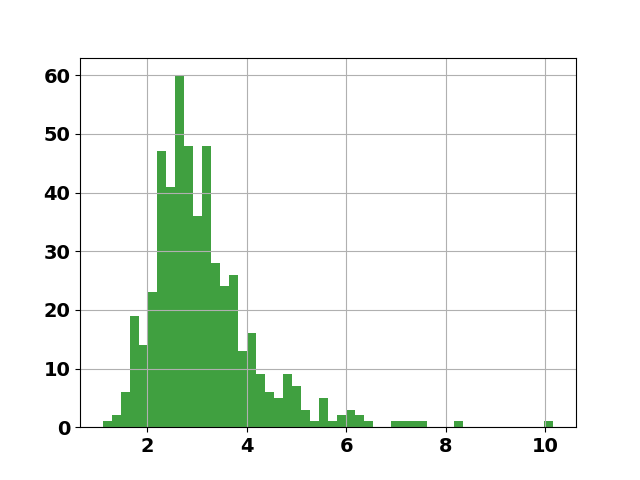}
    \end{subfigure}
    \hfill
    \begin{subfigure}[t]{0.24\textwidth}
    \centering
    \includegraphics[width=1.0\linewidth]{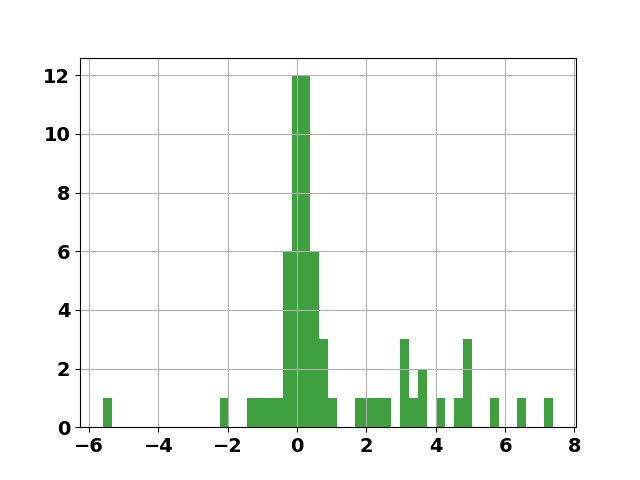}
    \end{subfigure}
    \hfill
    \begin{subfigure}[t]{0.24\textwidth}
    \centering
    \includegraphics[width=1.0\linewidth]{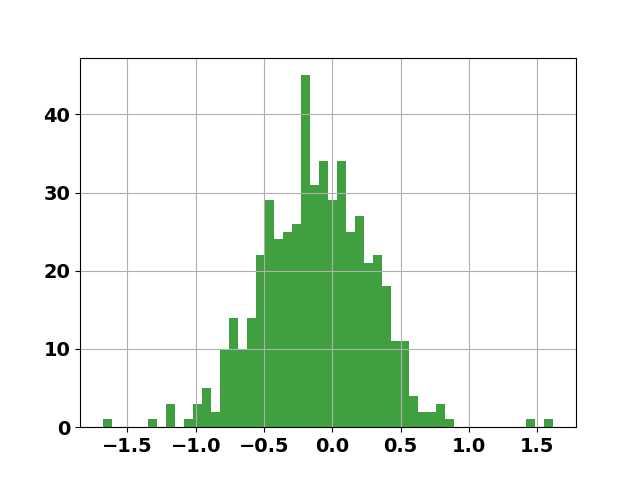}
    \end{subfigure}
    
    \caption{Distribution of the scaling factor $\myvector{\alpha}$ for various layers from the bottom to the top of the network (left to right) on a ResNet-18 trained until convergence on ImageNet. First row: $\myvector{\alpha}$ is computed using the analytical form proposed in~\cite{rastegari2016xnor}; Second row: $\myvector{\alpha}$ is computed using our proposed method (see Section~\ref{ssec:learnable-scaling}). Notice that our method allows for a more spread out distribution, that can take both positive and negative values, with significantly higher values that lead both to faster and more stable training.}
    \label{fig:alpha}
\end{figure*}

\subsubsection{Holistic Tucker decomposition}\label{sssec:holistic-tucker}

Motivated by the method proposed in~\cite{kossaifi2018parametrizing} and our finding from Section~\ref{sssec:layer-wise-tucker}, where we re-parametrized the weights using a layer-wise Tucker decomposition, herein we go one step further and propose to group together identically shaped weights inside the network in a higher-order tensor in order to exploit the inter relation between them holistically.

For ResNet-18~\cite{he2016deep} used for ImageNet classification, we create 3 groups of convolutional layers based on the macro-module structure characterizing the architecture. Each of these groups is then parameterized with a single 5-th order tensor $\mytensor{W'} \in  \mathbb{R}^{N \times O \times C \times w \times h}$ obtained by concatenating the weights of the $N$ convolutional layers in this group. The resulting decomposition is then defined as:
\begin{equation}
\text{sign}(\mytensor{W'}) = \text{sign}(\mytensor{G'} \times_0 \mymatrix{U}^{(0)} \times_1 \mymatrix{U}^{(1)} \times \cdots \times_4 \mymatrix{U}^{(4)}).
\label{eq:holistic}
\end{equation}
The individual weights of a given layer l can be obtained from $\mytensor{W} = \mytensor{W'}(l,:,:,:,:)$.

For the hourglass network used in our experiments for human pose estimation, we follow~\cite{kossaifi2018parametrizing} to derive  a \emph{single} 7-th  order tensor \mytensor{W}, the modes of which correspond to the number of HGs, the depth of each HG, the three signal pathways, the number of convolutional blocks, the number of input features, the number of output features, and finally the height and width of each of the convolutional kernels. The remaining few layers in the architecture are decomposed using a layer-wise Tucker decomposition. 

When tested on MPII, the proposed representation improves the performance with more than 3\% in terms of absolute error against the baseline and more than 1\% when compared with its layer-wise version (see Table~\ref{tab:main-results-mpii}). Similar results are observed on ImageNet (Table~\ref{tab:main-results-imagenet}).

\begin{table}[!htbp]
\small
	\begin{center}
		\begin{tabular}{|c|c|c|c|c|}
			\hline
			Decomposition & Holistic & Learn. alpha & Top-1 & Top-5 \\
			\hline\hline
			None & - & \xmark  & 52.3\% & 74.1\% \\
			None & - & \checkmark  & 53.0\% & 74.7\% \\
			SVD & \xmark & \checkmark  & 52.5\% & 74.2\% \\
			Tucker & \xmark & \xmark  & 54.0\% & 76.9\% \\
			Tucker & \xmark & \checkmark  & 54.7\%&  77.4\%\\
			Tucker & \checkmark & \xmark  & 55.2\% &  78.2\% \\
			\textbf{Tucker} & \checkmark & \checkmark  & \textbf{55.6}\% & \textbf{78.5}\% \\
            \hline
		\end{tabular}
	\end{center}
	\caption{Top-1 and Top-5 accuracy on the ImageNet dataset for different variations of the proposed binarization method.}
	\label{tab:main-results-imagenet}
\end{table}

\begin{figure*}[!htbp]
    \centering
    \begin{subfigure}[t]{0.23\textwidth}
    \centering
    \includegraphics[width=1.0\linewidth]{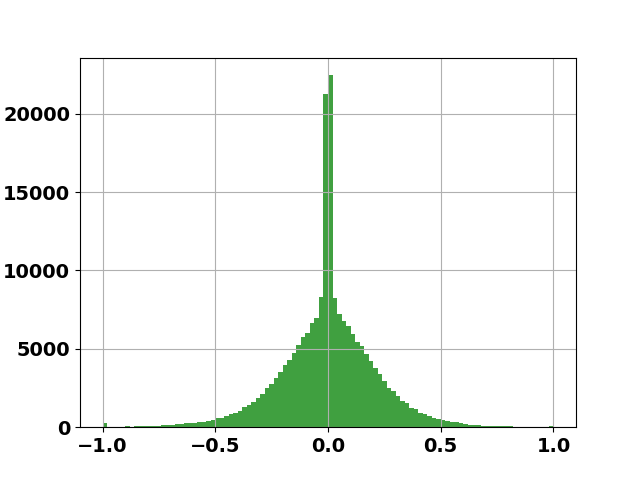}
    \end{subfigure}
    ~
    \begin{subfigure}[t]{0.23\textwidth}
    \centering
    \includegraphics[width=1.0\linewidth]{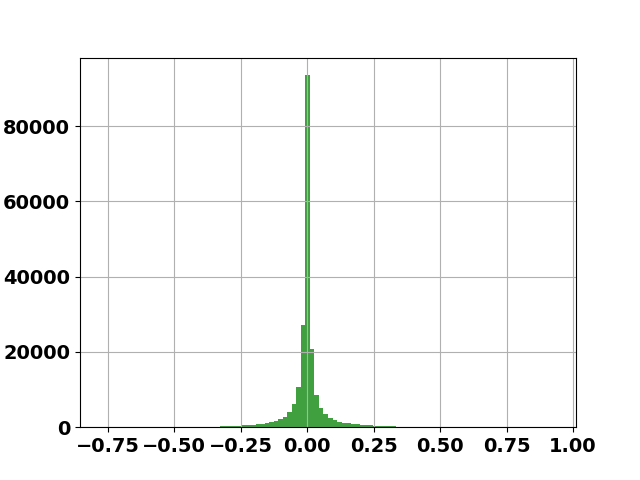}
    \end{subfigure}
    ~
    \begin{subfigure}[t]{0.23\textwidth}
    \centering
    \includegraphics[width=1.0\linewidth]{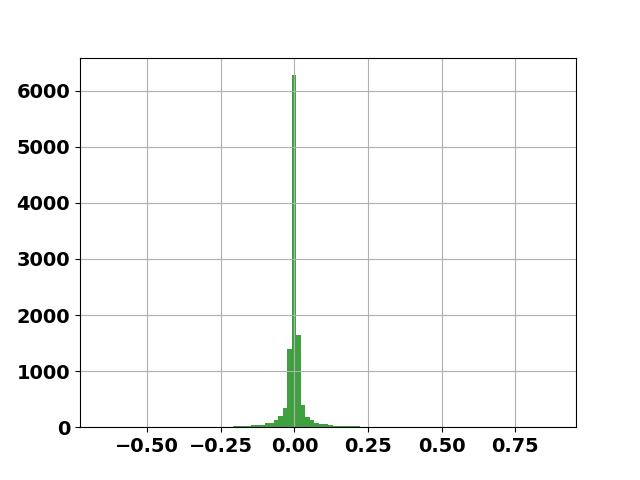}
    \end{subfigure}
    ~
    \begin{subfigure}[t]{0.23\textwidth}
    \centering
    \includegraphics[width=1.0\linewidth]{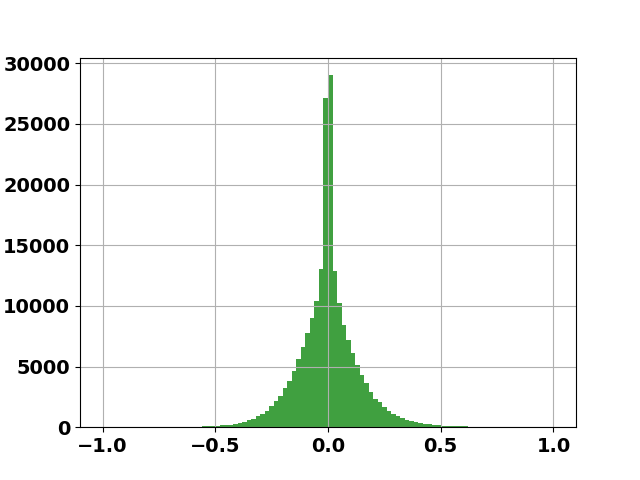}
    \end{subfigure}
    ~
    \begin{subfigure}[t]{0.23\textwidth}
    \centering
    \includegraphics[width=1.0\linewidth]{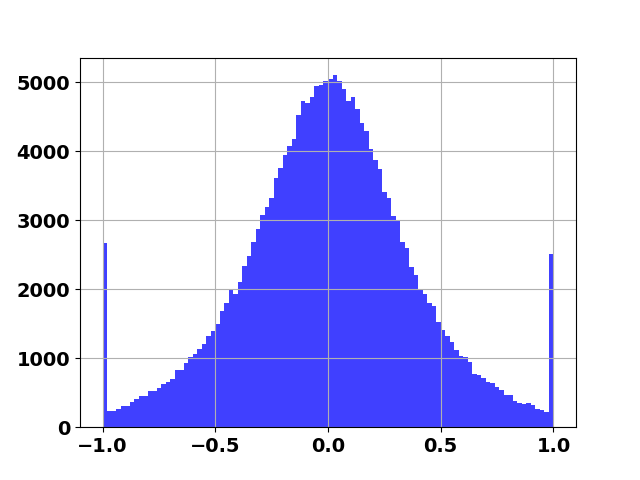}
    \end{subfigure}
     ~
    \begin{subfigure}[t]{0.23\textwidth}
    \centering
    \includegraphics[width=1.0\linewidth]{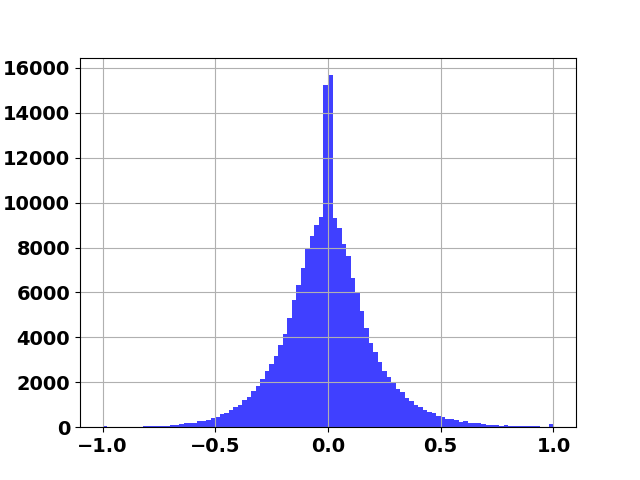}
    \end{subfigure}
    ~
    \begin{subfigure}[t]{0.23\textwidth}
    \centering
    \includegraphics[width=1.0\linewidth]{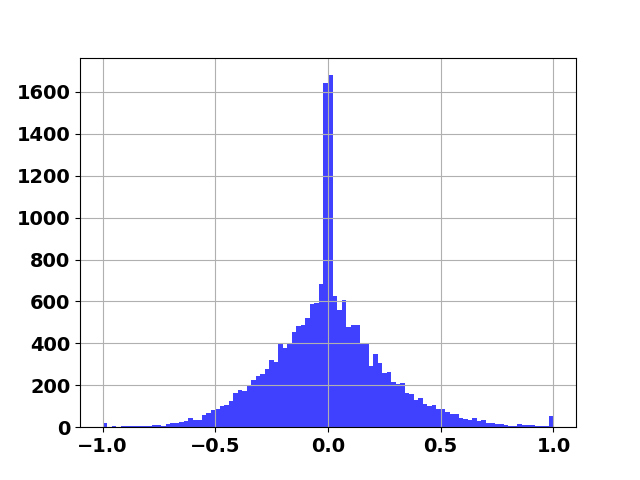}
    \end{subfigure}
    ~
    \begin{subfigure}[t]{0.23\textwidth}
    \centering
    \includegraphics[width=1.0\linewidth]{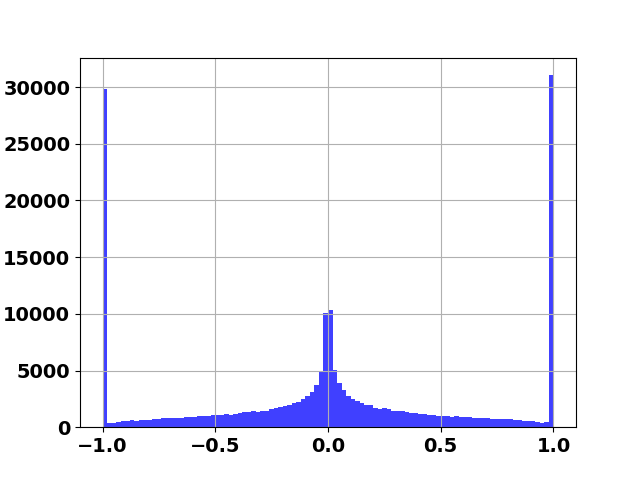}
    \end{subfigure}
    
    \caption{Distribution of the weights before binarizing them using the $\textit{sign}$ function for various layers from the bottom to the top of the network (left to right) on a HourGlass trained on MPII. First row: the weights are obtained using no parameterization method (i.e using the method from~\cite{rastegari2016xnor,bulat2017binarized}) Second row: the weights are computed using our proposed method and reconstructed using a holistic Tucker parameterization (see Section~\ref{sssec:holistic-tucker}).}
    \label{fig:weights}
\end{figure*}

\subsection{Learnable scaling factors}\label{ssec:learnable-scaling}

One of the key ingredients of the recent success of binarized neural network was the introduction of the $\myvector{\alpha}$ weight scaling factor in~\cite{rastegari2016xnor} (see Eq.~\ref{eq:binarize}), computed analytically as the average of absolute weight values. While this estimation generally performs well, it attempts to minimize the difference between the real weights and the binary ones $\mytensor{W} \approx \myvector{\alpha} \text{sign}(\mytensor{W})$ and does not explicitly decrease the overall network loss. In contrast, in this work we propose to learn the scaling factor by minimizing its value with respect to the networks cost function, learning it discriminatively via back-propagation.

Fig.~\ref{fig:alpha} shows the difference between the scaling factors learned using our proposed method vs the ones computed using the analytic solution from~\cite{rastegari2016xnor}. Note that our method leads to (a) a more spread out distribution that can take both positive and negative values, (b) has significantly higher magnitude, thus leading to a faster and more stable training.

Table~\ref{tab:main-results-mpii} and~\ref{tab:main-results-imagenet} show that the newly proposed method for learning the scale factor offers consistent gains across all decompositions and tasks (both human pose estimation and image recognition), with the largest gain observed for the MPII dataset (more than 1\%).
 \section{Experimental evaluation}\label{sec:experiments}

\begin{figure}
    \centering
    \includegraphics[width=0.9\linewidth,trim={0.5cm 0.5cm 0.5cm 0.5cm},clip]{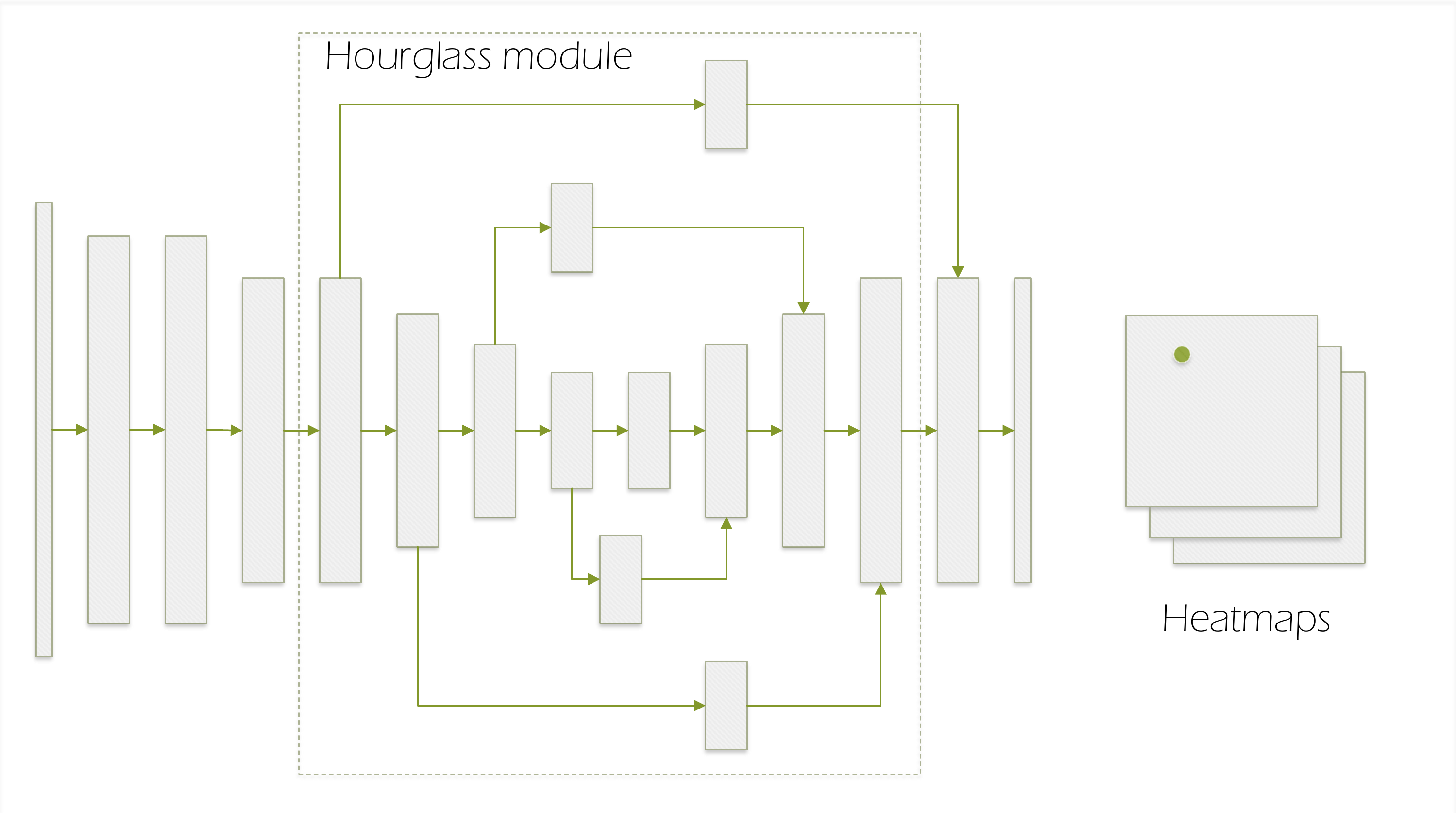}
    \caption{The hourglass architecture as introduced in~\cite{newell2016stacked} using binarized basic blocks as building modules.}
    \label{fig:hg-arch}
\end{figure}

This section firstly presents the experimental setup, network architecture and training procedure. We then empirically demonstrate the advantage of our approach on single person human pose estimation and large-scale image recognition where we surpass the state-of-the-art by more than 4\% (Section ~\ref{ssec:sota}).

\subsection{Human pose estimation}

\textbf{Datasets.} MPII~\cite{andriluka20142d} is one of the most challenging human pose estimation datasets to-date consisting of over 40,000 people, each annotated with up to 16 keypoints and visibility labels. The images were extracted from various youtube videos. For training/validation split, we used the same partitioning as introduced in~\cite{tompson2014joint}. The results are reported in terms of PCKh~\cite{andriluka20142d}.

\begin{figure*}
    \centering
    \includegraphics[width=0.9\linewidth,trim={0.5cm 0.5cm 0.5cm 0.5cm},clip]{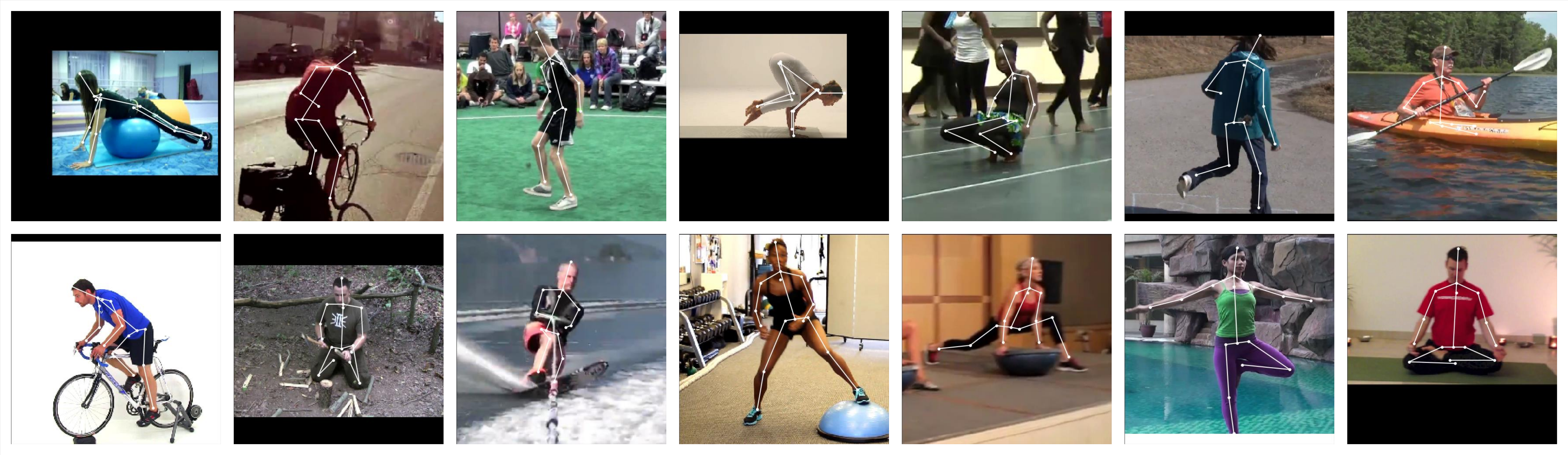}
    \caption{Qualitative examples produced by our binary method on the validation set of MPII. Notice that our method can cover a large variety of poses and across a large number of different activities.}
    \label{fig:examples}
\end{figure*}

\textbf{Network architecture.} The Hourglass (HG)~\cite{newell2016stacked} and its variants represent the current state-of-the-art on human pose estimation. As such, in this work, we used an hourglass-like architecture (Fig.~\ref{fig:hg-arch}) constructed using the basic blocks introduced in~\cite{he2016deep, rastegari2016xnor} (see also Fig.~\ref{fig:basic-block}). The HG network as a whole follows an encoder-decoder structure with skip connections between each corresponding level of the decoder and encoder part. The basic block used has 128 channels.

\noindent\textbf{Training.} During training, we followed the best practices and randomly augmented  the  data  with  rotation  (between  $-30^\circ$ and $30^\circ$ degrees), flipping and scale jittering (between 0.7 and 1.3). All models were trained until convergence (typically 120 epochs max). During this time, the learning rate was dropped multiple times from $2.5e-4$  to $5e-6$. We used no weight decay.

All of our models were trained using pytorch~\cite{paszke2017automatic} and RMSProp~\cite{tieleman2012lecture}. The tensor operations were implemented using tensorly~\cite{kossaifi2016tensorly}.

\subsection{Large-scale image classification}

\textbf{Datasets} ImageNet~\cite{deng2009imagenet} is a large scale image recognition dataset consisting of more than 1.2M images for training distributed over 1000 object classes and 50,000 images for validation.

\noindent\textbf{Network architecture.} Following~\cite{rastegari2016xnor,courbariaux2015binaryconnect}, we used a Resnet-18~\cite{he2016deep} architecture for our experiments on ImageNet. The ResNet-18 consists of 18 convolutional layers distributed across 4 macro-modules that are linked via a skip-connection. At the beginning of each macro-module the resolution is dropped using a convolutional layer with a stride $>1$. The final predictions are obtained by using an average pooling layer followed by a fully connected one.

\noindent\textbf{Training.} During training, we resized the input images to $256\times 256$px and then a random $224\times224$px crop was selected for training. At test time, instead of random cropping the images, a center crop was applied. The network was trained using Adam~\cite{kingma2014adam} for 90 epochs with a learning rate of $1e-3$ that was gradually reduced (dropped every 30 epochs) to $1e-6$. The weight decay was set to $1e-7$ for the entire duration of the training.

\subsection{Comparison with state-of-the-art}\label{ssec:sota}

\begin{table}[!htbp]
	\begin{center}
    \small{
		\begin{tabular}{|c|c|c|}
			\hline
			Method & \#parameters & PCKh \\
			\hline\hline
			HBC~\cite{bulat2017binarized} & 6.2M  & 78.1\% \\
			\hline
			\textbf{Ours} &  6.0M  & \textbf{82.5}\% \\
			\hline
			Real valued & 6.0M & 85.8\% \\
			\hline
		\end{tabular}
		}
	\end{center}
	\caption{Comparison with the state-of-the-art method of~\cite{bulat2017binarized} on the validation set of the MPII dataset. Our method improves upon the state-of-the-art approach of~\cite{bulat2017binarized} by mote than 3\% further bridging the gap between the real and binary domain.}
	\label{tab:results-sota-mpii}
\end{table}

In this section, we report the performance of our method on the challenging and diverse tasks of human pose estimation (on MPII) and large scale-image recognition (on Imagenet), and compare it with that of published  state-of-the-art methods that use fully binarized neural networks (i.e both the weights and the features are binary). 

On human pose estimation, the only other work that trains fully binarized networks is that of~\cite{bulat2017binarized}. As the results from Table~\ref{tab:results-sota-mpii} show, our method offers an improvement of more than 4\% on the MPII dataset when compared against the state-of-the-art method of~\cite{bulat2017binarized}. Qualitative results are shown in figure~\ref{fig:examples}.

As Table~\ref{tab:results-sota-imagenet} shows, for ImageNet classification, our method improves upon the results from~\cite{rastegari2016xnor} by up to 5\% in terms of absolute error.

\begin{table}[!htbp]
\small{
	\begin{center}
		\begin{tabular}{|c|c|c|}
			\hline
			Method & Top-1 accuracy & Top-5 accuracy \\
			\hline\hline
			BNN~\cite{courbariaux2016binarized} & 42.2\% & 69.2\% \\
			XNOR-Net~\cite{rastegari2016xnor} & 51.2\% & 73.2\% \\
			\hline
			\textbf{Ours} & \textbf{55.6}\%  & \textbf{78.5}\% \\
			\hline
			Real valued~\cite{he2016deep} & 69.3\% & 89.2\% \\
            \hline
		\end{tabular}
	\end{center}
	}
	\caption{Top-1 and Top-5 accuracy on ImageNet using a ResNet-18 binarized architecture. Notice that out methods surpass the current state-of-the-art by a large margin, up to 5\% improvement in terms of absolute error.}
	\label{tab:results-sota-imagenet}
\end{table} 
\section{Conclusion}\label{sec:conclusion}

In this paper, we proposed a novel binarization method in which the weight tensor of each layer or group of layers is parametrized using matrix or tensor decomposition. The binarization process is then performed using this latent parametrization, via a quantization function (e.g. sign function) applied to the reconstructed weights. 

This simple idea enforces a coupling of the filters before binarization which is shown to significantly improve the accuracy of the trained models. Additionally, instead of computing the weight scaling factor analytically we propose to learn them via backpropagation. When evaluated on single person human pose estimation (on MPII) and large scale image recognition (Imagenet) our method surpasses the state-of-the-art by 4\%, and respectively 5\% while retaining the speed-up (up to $58\times$) and space saving (up to $32\times$) typically offered by binary networks.
 
{\small
\bibliographystyle{ieee}

}

\end{document}